\documentclass{article}

\usepackage[preprint,nonatbib]{neurips_2025}

\usepackage[utf8]{inputenc} 
\usepackage[T1]{fontenc}    
\usepackage{url}            
\usepackage{booktabs}       
\usepackage{amsfonts}       
\usepackage{nicefrac}       
\usepackage{microtype}      
\usepackage{xcolor}         

\usepackage{graphicx}
\usepackage{amsmath}
\usepackage{ulem}
\definecolor{lightblue}{RGB}{0, 110, 204}
\usepackage[pagebackref=true,breaklinks=true,letterpaper=true,citecolor=lightblue,colorlinks,bookmarks=false]{hyperref}
\usepackage{colortbl}
\usepackage{titlesec}
\titlespacing*{\paragraph}{0pt}{0ex}{0.5em}

\title{AnyCharV: Bootstrap Controllable Character Video Generation with Fine-to-Coarse Guidance}

\author{
Zhao Wang$^1$\footnotemark[1],
Hao Wen$^2$\footnotemark[1],
Lingting Zhu$^3$,
Chenming Shang$^2$,
Yujiu Yang$^2$\footnotemark[2],
Qi Dou$^1$\footnotemark[2] \\
$^1$The Chinese University of Hong Kong,
$^2$Tsinghua University,
$^3$The University of Hong Kong \\
\texttt{\{zwang21@cse., qidou\}@cuhk.edu.hk, ltzhu99@connect.hku.hk} \\
\texttt{\{wenh22@mails., scm22@mails., yang.yujiu@sz.\}tsinghua.edu.cn} \\
\vspace{2mm}
\texttt{\url{https://anycharv.github.io}}
}

\begin{document}

\maketitle

\begin{figure}[!htb]
  \centering
  \vspace{-6mm}
   \includegraphics[width=\linewidth]{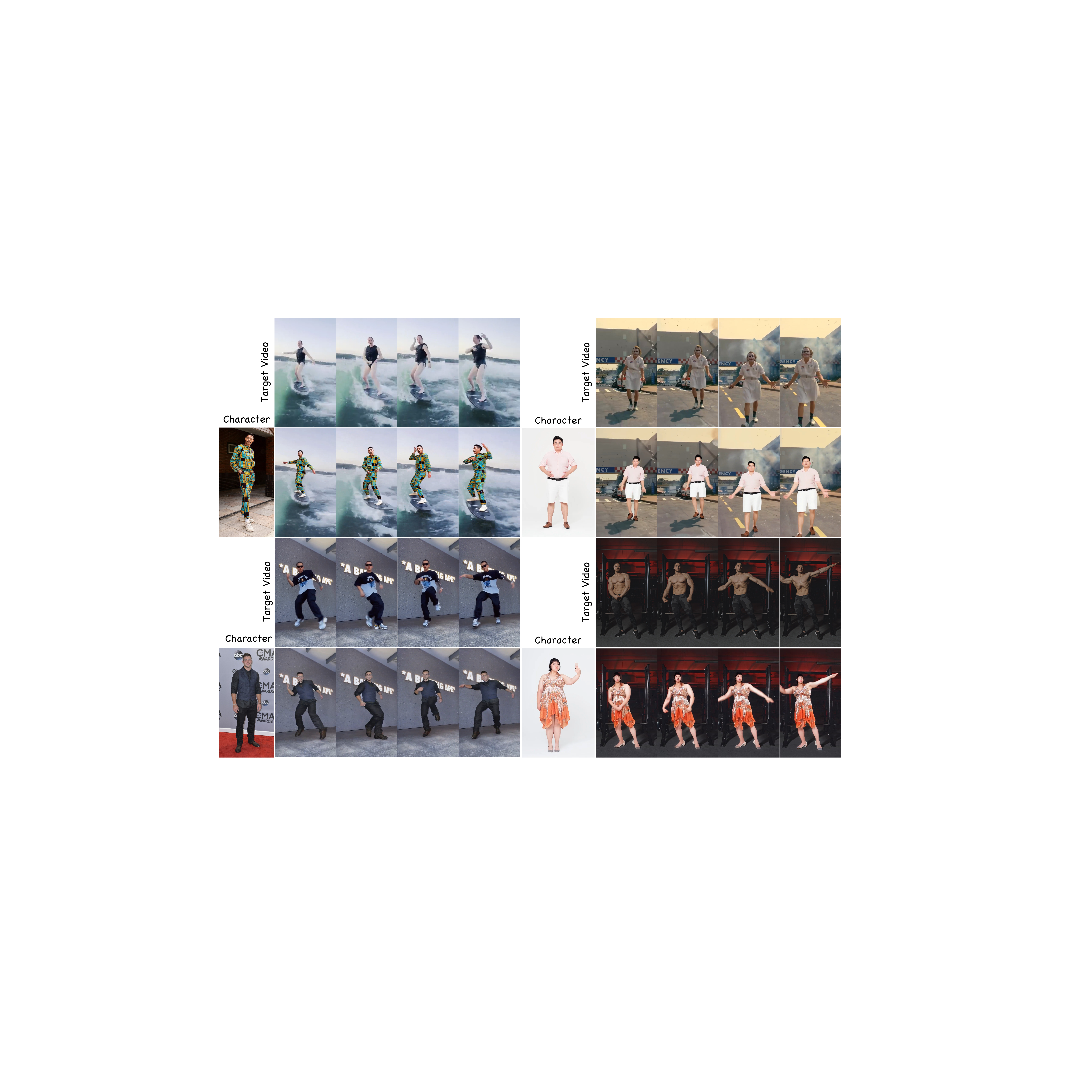}
   \vspace{-6mm}
   \caption{
   Our method works by giving a reference character image (left) and a target driving video (top) supplying complex motion, background scene, and interaction information. Our method can naturally synthesize the given arbitrary reference character following the motion in the target video, preserving the real-world scene and complex interaction.
   }
   \label{fig:teaser}
\end{figure}

\renewcommand{\thefootnote}{\fnsymbol{footnote}} 
\footnotetext[1]{Equal contributions.}
\footnotetext[2]{Corresponding authors.}
\renewcommand{\thefootnote}{\arabic{footnote}}

\begin{abstract}
Character video generation is a significant real-world application focused on producing high-quality videos featuring specific characters. Recent advancements have introduced various control signals to animate static characters, successfully enhancing control over the generation process. However, these methods often lack flexibility, limiting their applicability and making it challenging for users to synthesize a source character into a desired target scene. To address this issue, we propose a novel framework, \textit{\textbf{AnyCharV}}, that flexibly generates character videos using arbitrary source characters and target scenes, guided by pose information. Our approach involves a two-stage training process. In the first stage, we develop a base model capable of integrating the source character with the target scene using pose guidance. The second stage further bootstraps controllable generation through a self-boosting mechanism, where we use the generated video in the first stage and replace the fine mask with the coarse one, enabling training outcomes with better preservation of character details. Extensive experimental results demonstrate the superiority of our method compared with previous state-of-the-art methods.
\end{abstract}

\section{Introduction}


Recent advancements in diffusion modeling have significantly advanced character video generation for various applications \cite{blattmann2023stable, ho2022video, hong2022cogvideo, wu2023tune, zhang2023i2vgen, wang2024customvideo, wu2024motionbooth,zhao2025motiondirector}. While these approaches achieve remarkable success with guidance from text and image priors, they often fail to preserve the intrinsic characteristics of the character, such as nuanced appearance and complex motion. Additionally, these methods lack the capability for users to exert individual control during the character video generation process, making it difficult to modify the character's motion and the surrounding scene. These limitations hinder the practical application and development of character video generation technologies.

Several recent works aim to improve the controllability of character video generation by incorporating additional conditions, such as pose \cite{hu2024animate, xu2024magicanimate, zhang2024mimicmotion, zhu2025champ}, depth \cite{xing2025make, zhu2025champ}, and normal maps \cite{zhu2025champ}. 
However, these methods primarily focus on animating characters within fixed background scenes or customizing characters against random backgrounds, lacking the capability for precise background control and often introducing artifacts. 
These limitations reduce the flexibility of character video generation and imposes significant constraints on practical applications. 
Directly replacing a character within a given scene is crucial for applications such as art creation and movie production. 
Some related works attempt to address this issue using 3D priors \cite{men2024mimo, viggle2024} and naive 2D representations \cite{zhao2023make, qiu2024moviecharacter,jiang2025vace}. However, characters constructed from 3D information often lack realistic interaction appearances, while 2D prior-driven characters suffer from temporal inconsistencies and poor identity preservation.

In this work, we aim to develop a flexible and high-fidelity framework for controllable character video generation. 
Given the complexity and challenges of this task, we propose a novel two-stage approach with fine-to-coarse guidance, \textbf{\textit{AnyCharV}}, consisting of a self-supervised composition stage and a self-boosting training stage.
In the first stage, we construct a base model to integrate the reference character and the target scene, using fine guidance from the target character's segmentation mask and the pose as a conditional signal. 
During this phase, the model learns to accurately compose the source character and target scene with precise spatial and temporal control.
However, due to shape differences between the source and target characters, the generated video often appears blurry, contains artifacts, and lacks realism.
To address this critical issue, which is a significant drawback of current research, we introduce a self-boosting training stage. 
In this stage, we generate multiple source-target video pairs using the model from the first stage and employ these synthesized videos for bootstrap training. 
These video pairs share the same background scene, with the source and target characters in identical poses. 
Instead of using a fine target character segmentation mask, we utilize a bounding box mask to provide coarse guidance for the character area. 
By focusing solely on the source character and not separately feeding the model the target video scene, we enable better character control and detail recovery during self-boosting training. 
This approach ensures that the details of the source character are better preserved during subsequent inference, as shown in Figure \ref{fig:teaser}.
In summary, our contributions are as follows:
\begin{itemize}
    \item We propose a novel framework AnyCharV for controllable character video generation that employs fine-to-coarse guidance. This approach enables the seamless integration of any source character into a target video scene, delivering flexible and high-fidelity results.
    \item We achieve the composition of the source character and target video scene in a self-supervised manner under the guidance of a fine segmentation mask, introducing a simple yet effective method for controllable generation.
    \item We develop a self-boosting strategy that leverages the interaction between the target and source characters with guidance from a coarse bounding box mask, enhancing the preservation of the source character's details during inference.
    \item Our method demonstrates superior generation results, outperforming previous open-sourced and close-sourced state-of-the-art (SOTA) approach both qualitatively and quantitatively.
\end{itemize}


\section{Related Work}

\subsection{Controllable Video Generation}
Controlling the process of video generation is crucial and has been extensively studied. 
Recent research efforts in this direction often rely on introducing additional control signals, such as depth maps \cite{chai2023stablevideo, zhang2023controlvideo, wang2024videocomposer}, canny edges \cite{zhang2023controlvideo}, text descriptions \cite{zhang2023i2vgen}, sketch maps \cite{wang2024videocomposer}, and motions \cite{wang2024videocomposer}. 
These works achieve control by incorporating these signals into the video generation model, resulting in the desired outcomes.
Meanwhile, other approaches focus on learning high-level feature representations related to appearance \cite{he2024id, wang2024customvideo, wei2024dreamvideo} and motion \cite{wu2024motionbooth, yang2024direct, zhao2025motiondirector} within video diffusion models. 
These methods aim to customize the video generation by learning the identity of subjects and motions.
However, the controls within these approaches are too coarse to meet the stringent requirements for controllable character video generation, where fine-grained control over both character and scene is essential.


Some recent works trying to offer more precise control and shape retargeting to improve the generation quality.
Zhang et al. \cite{zhang2023skinned} proposes a residual retargeting network to adjust the source motions to fit the target skeletons and shapes progressively. MeshRet \cite{ye2024skinned} further improves the retargeting performance via introducing the body geometries. STAR \cite{chai2024star} and SMT \cite{zhang2024semantics} incorporate the text descriptions, leveraging the text-to-motion and vision-language models, to help handle the shape differences during animation. However, these works require explicit 3D representations, e.g., meshes, which is expensive to be created via hand-crafters and difficult to be reconstructed with high-quality from videos. On the contrary, our method focuses on the simplest creation setting where only the character and the reference video is supplied, easing the creation while achieving impressive results.






\subsection{Character Video Generation}

Character video generation has achieved remarkable results recently with the development of diffusion models \cite{ho2020denoising,song2020denoising,lu2022dpm}.
Generating a character video with high-fidelity is challenging due to the complex character motion and various appearance details.
To achieve high-quality generation, some approaches try to train a large model upon vast amounts of data \cite{liu2024sora,yang2024cogvideox,li2024openhumanvid,guo2023animatediff,blattmann2023stable,bao2024vidu}.
In this case, the character video can be directly synthesized by the input text prompt or reference image.
Nevertheless, these direct methods introduce significant randomness when generating the characters, resulting in many artifacts and unsmooth motions.
To tackle these issues, recent researches try to guide the character video generation with 2D pose sequence \cite{hu2024animate,xu2024magicanimate,peng2024controlnext,chang2023magicdance,chang2024magicpose,zhai2024idol,wang2024disco}, depth maps \cite{zhu2025champ}, and 3D normal \cite{zhu2025champ}.
However, these approaches are limited to animating the given image within its original background scene, which restricts their applicability in diverse scenarios.

Recent works have proposed more flexible settings by combining arbitrary characters with arbitrary background scenes. An early work, Make-A-Protagonist \cite{zhao2023make}, modifies the foreground character using image prompts and mask supervision, while suffers from poor temporal consistency and unexpected jittering. 
MIMO \cite{men2024mimo} addresses this by decoupling the foreground character and background scene for further composition during video generation. 
Although MIMO's approach is based on 3D representations, it can result in unrealistic video generation. 
Additionally, inaccuracies in conditioned 3D motion, occlusion, and structure modeling can degrade performance. 
Viggle \cite{viggle2024}, which is also based on 3D representations, encounters similar challenges.
MovieCharacter \cite{qiu2024moviecharacter} introduces a tuning-free framework for video composition using pose-guided character animation and video harmonization \cite{guerreiro2023pct}. 
However, MovieCharacter struggles with complex motions and interactions. 
VACE \cite{jiang2025vace} introduce a inpainting-based model for swapping the character in the target video, but falls back in identity preservation.
In contrast to these methods, we address the composition problem using a self-supervised training framework with fine-to-coarse guidance driven by 2D pose guidance, enhancing composition capabilities and improving generation quality.


\begin{figure*}[t]
    \centering
    \includegraphics[width=\linewidth]{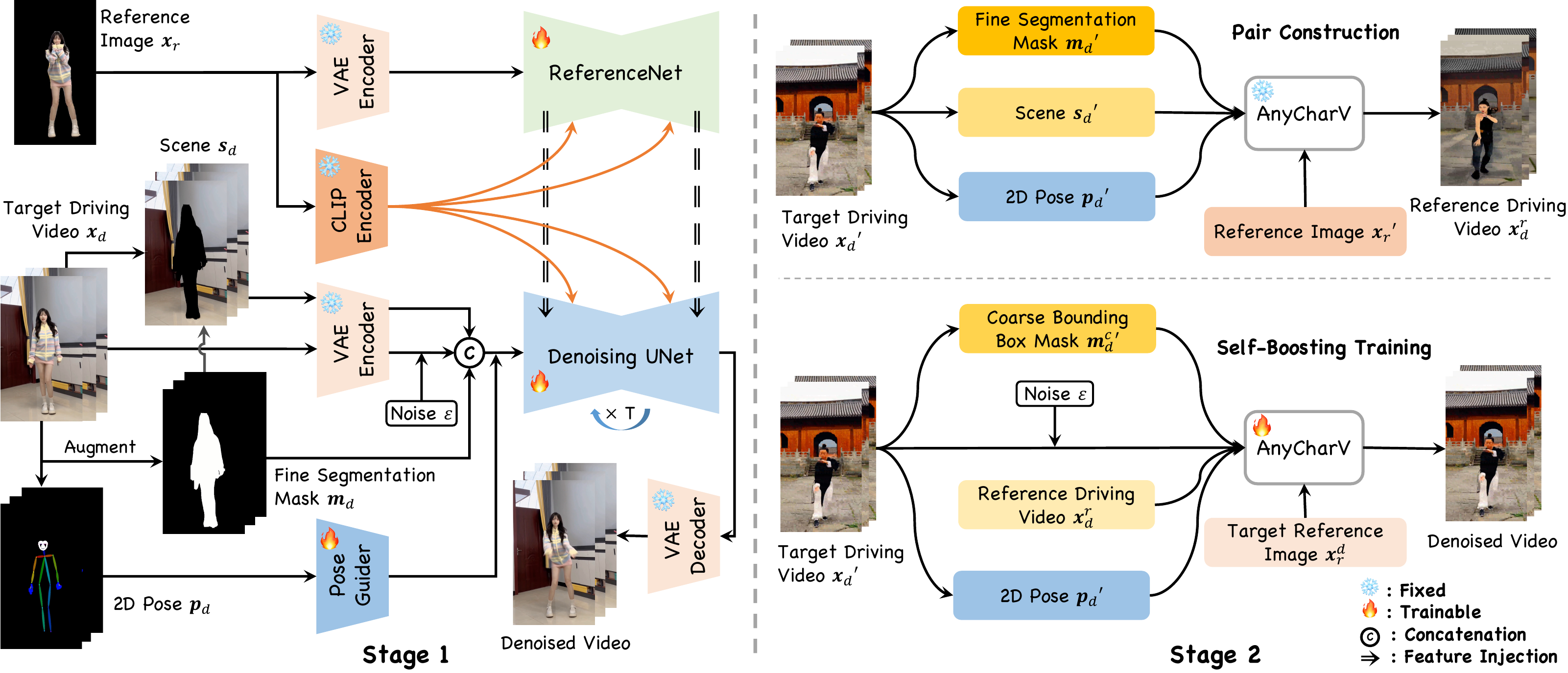}
    \vspace{-7mm}
    \caption{The overview of our proposed AnyCharV. We design a two-stage pipeline with fine-to-coarse guidance for controllable character video generation. In the first stage, we utilize a self-supervised manner to train a base model for composing a reference character with the target scene $\boldsymbol{s}_d$, guided by fine segmentation mask $\boldsymbol{m}_d$ and 2D pose sequence $\boldsymbol{p}_d$. In the second stage, we propose a self-boosting training strategy by interacting between the reference and target character using coarse bounding box mask guidance. The CLIP encoder and VAE are always frozen. We train denoising UNet, ReferenceNet, and pose guider during the first stage, while only finetuning denoising UNet in the second stage.}
    \label{fig:framework}
\end{figure*}

\section{Method}

We aim to generate a video featuring our desired character, guided by a target video that provides character motion, background scene, and interaction information. This task presents several significant challenges.
Firstly, the appearance of the given character must be well-preserved.
Secondly, the model should accurately animate the character to follow the target motion.
Thirdly, the model is required to ensure the seamless integration of the generated character into the target scene within the corresponding region.
To address these challenges, we design a two-stage training framework \textbf{\textit{AnyCharV}} with fine-to-coarse guidance.
In the following, we successively describe the preliminary of video diffusion model in Section \ref{sec:preliminary}, the first stage self-supervised composition with fine mask guidance in Section \ref{sec:composition}, and the second stage self-boosting training strategy with coarse mask guidance in Section \ref{sec:self_boost}.
An overview of our proposed framework is illustrated in Figure \ref{fig:framework}.


\subsection{Preliminary: Video Diffusion Model}
\label{sec:preliminary}

Video diffusion models (VDMs)~\cite{chen2024videocrafter2,blattmann2023stable,yang2024cogvideox} synthesize videos by successively refining randomly sampled Gaussian noise $\epsilon$. 
This procedure parallels the reversal of a fixed-length Markov chain. 
Through iterative denoising, VDMs capture the temporal relationships embedded in video data. 
Concretely, at each timestep $t \in \{1, 2, \dots, T\}$, a video diffusion model $\Theta$ estimates the noise given an image condition $\boldsymbol{c}_{img}$. 
The training process optimizes a reconstruction objective:
\begin{equation}
\mathcal{L}=\mathbb{E}_{\epsilon, \boldsymbol{z}, \boldsymbol{c}_{img}, t}\left[\left\|\epsilon-\epsilon_{\Theta}\left(\boldsymbol{z}_t, \boldsymbol{c}_{img}, t\right)\right\|_2^2\right],
\label{equ:vdm}
\end{equation}
where $\boldsymbol{z} \in \mathbb{R}^{B \times L \times H \times W \times D}$ denotes the latent representation of the video, with $B$ as the batch size, $L$ the sequence length, $H$ and $W$ the spatial dimensions, and $D$ the latent dimensionality. 
Here, $\epsilon_{\Theta}$ is the model-estimated noise.
The noisy intermediate state $\boldsymbol{z}_{t}$ is obtained by combining ground-truth features $\boldsymbol{z}_{0}$ with noise $\epsilon$ as $\boldsymbol{z}_{t} = \alpha_{t} \boldsymbol{z}_{0} + \sqrt{1 - \alpha_{t}^{2}} \, \epsilon$, where $\alpha_{t}$ is a diffusion parameter.

\subsection{Self-Supervised Composition with Fine Guidance}
\label{sec:composition}

\paragraph{Naive Self-Supervised Strategy.}
To train our model, it is impractical to collect a large number of video pairs that feature the same character motion and scene but with different characters. 
Therefore, we develop a naive self-supervised strategy to make our training feasible.
Given a video sample $\boldsymbol{X}$, we sample a reference image $\boldsymbol{x}_r$ and a target driving video $\boldsymbol{x}_d$.
We extract the corresponding character segmentation mask $\boldsymbol{m}_d$ from $\boldsymbol{x}_d$ by a segmentation model, such as SAM2 \cite{ravi2024sam}.
Then, we can train the model to generate $\boldsymbol{x}_d$ with the reference character $\boldsymbol{x}_r$ and the target scene $\boldsymbol{s}_d = \overline{\boldsymbol{m}}_d \odot \boldsymbol{x}_d$, where $\overline{\boldsymbol{m}}_d=\boldsymbol{J}-\boldsymbol{m}_d$ is the complementary part of the mask $\boldsymbol{m}_d$, and $\boldsymbol{J}$ is a all-ones matrix.


\paragraph{Reference Identity Preservation.}
To preserve the appearance and identity of the reference character, we utilize the reference character via two pathways.
Firstly, the reference image $\boldsymbol{x}_r$ is encoded by a pre-trained CLIP \cite{radford2021learning} image encoder and conditioned on the cross-attention layers.
Secondly, with the success of previous character animation works \cite{hu2024animate,xu2024magicanimate}, we utilize a ReferenceNet to better encode the character appearance.
The spatial features from ReferenceNet are concatenated with spatial features from denoising UNet in the spatial attention layers for self-attention modeling.
We remove the background scene in the reference image to avoid introducing redundant context information.


\paragraph{Fine Mask Supervision.}

As mentioned before, our model should achieve the seamless integration for the reference character $\boldsymbol{x}_r$ and the target scene $\boldsymbol{s}_d$.
Naively only taking the reference character and the target scene will lead to high randomness of the composition process.
To this end, we control the generation by introducing the target character segmentation mask $\boldsymbol{m}_d$ to guide the generation region.
This fine segmentation mask is directly concatenated on the noisy target video together with the target scene as a whole.
However, we find that such fine mask will introduce some harmful shape information, especially at the border of the fine mask (see Figure \ref{fig:component} 'w/o Self-Boost').
To reduce this bad effect, we apply augmentation onto the input fine segmentation mask $\boldsymbol{m}_d$ to make it have irregular mask border.
And the scene is obtained by masking $\boldsymbol{x}_d$ with the augmented mask $\boldsymbol{m}_d$.

\paragraph{Pose Guidance.}
Recent works \cite{hu2024animate,xu2024magicanimate} has achieved great success on animating arbitrary character with arbitrary pose sequence.
Similar with them, we build a lightweight pose guider to embed the target character pose $\boldsymbol{p}_d$, consisting of 4 convolution layers, and the embedded pose images are added to the noisy target video latents.
The pose $\boldsymbol{p}_d$ can be extracted from the target video $\boldsymbol{x}_d$ via a pose detector, such as DWPose \cite{yang2023effective}.


\subsection{Self-Boosting Training with Coarse Guidance}
\label{sec:self_boost}

The based model developed in Section \ref{sec:composition} suffers from the bad shape effect caused by the border of fine segmentation mask.
To address this, we introduce a self-boosting strategy to mitigate the negative impact of the fine mask shape. 
This strategy aims to replace the fine segmentation mask with the coarse bounding box mask to indicate the approximate region. 
Accordingly, the target scene is provided with the original video instead of the masked video to avoid introduce excessive shape borders, while promoting the direct interaction between the reference and target characters.
Then, we will show how to achieve this training process in detail.

\paragraph{Pair Construction.}

As mentioned in the beginning of Section \ref{sec:composition}, we lack video pairs with the same character motion and scene but different characters to train our model.
Given a base model developed in Section \ref{sec:composition}, it becomes possible to create such pairs through generation.
We create the data pairs via drawing samples in the original dataset, where we choose one video for reference character and another for driving video.
Specifically, we randomly sample a frame from one video to serve as the reference character image $\boldsymbol{x}_r'$ and regard the other video as the target driving video $\boldsymbol{x}_d'$.
Given $\boldsymbol{x}_r'$ and $\boldsymbol{x}_d'$, we can generate a result video, where we name it as reference driving video $\boldsymbol{x}_d^r$.
To this end, we obtain a video pair $(\boldsymbol{x}_d^r, \boldsymbol{x}_d')$, in which the character motion, scene, and interaction are all the same while they features different characters.
We build 64,000 video pairs following the above process.




\paragraph{Self-Boosting Training.}

With this paired video dataset constructed, it comes possible for further boosting the base model with original unmasked video and coarse bounding box mask.
During training, instead of separately extracting the scene from the target driving video $\boldsymbol{x}_d'$, we take the generated reference driving video $\boldsymbol{x}_d^r$ as the input to provide the scene information.
To help the model localize the corresponding area of the target character, we provide a coarse bounding box mask $\boldsymbol{m}_d^{c\prime}$ as the coarse guidance.
Given the reference driving video $\boldsymbol{x}_d^r$ and coarse bounding box mask $\boldsymbol{m}_d^{c\prime}$, the model learns to build interactions between the reference and target characters, as well as integrate the reference character with the background scene by implicitly learning the interactions in the latent space.
By loosening the mask guidance from fine to coarse, we finally mitigate the negative influence of the fine mask shape, resulting in more natural generated videos that better preserve the details of the reference character.


\begin{figure*}[!t]
    \centering
    \includegraphics[width=\linewidth]{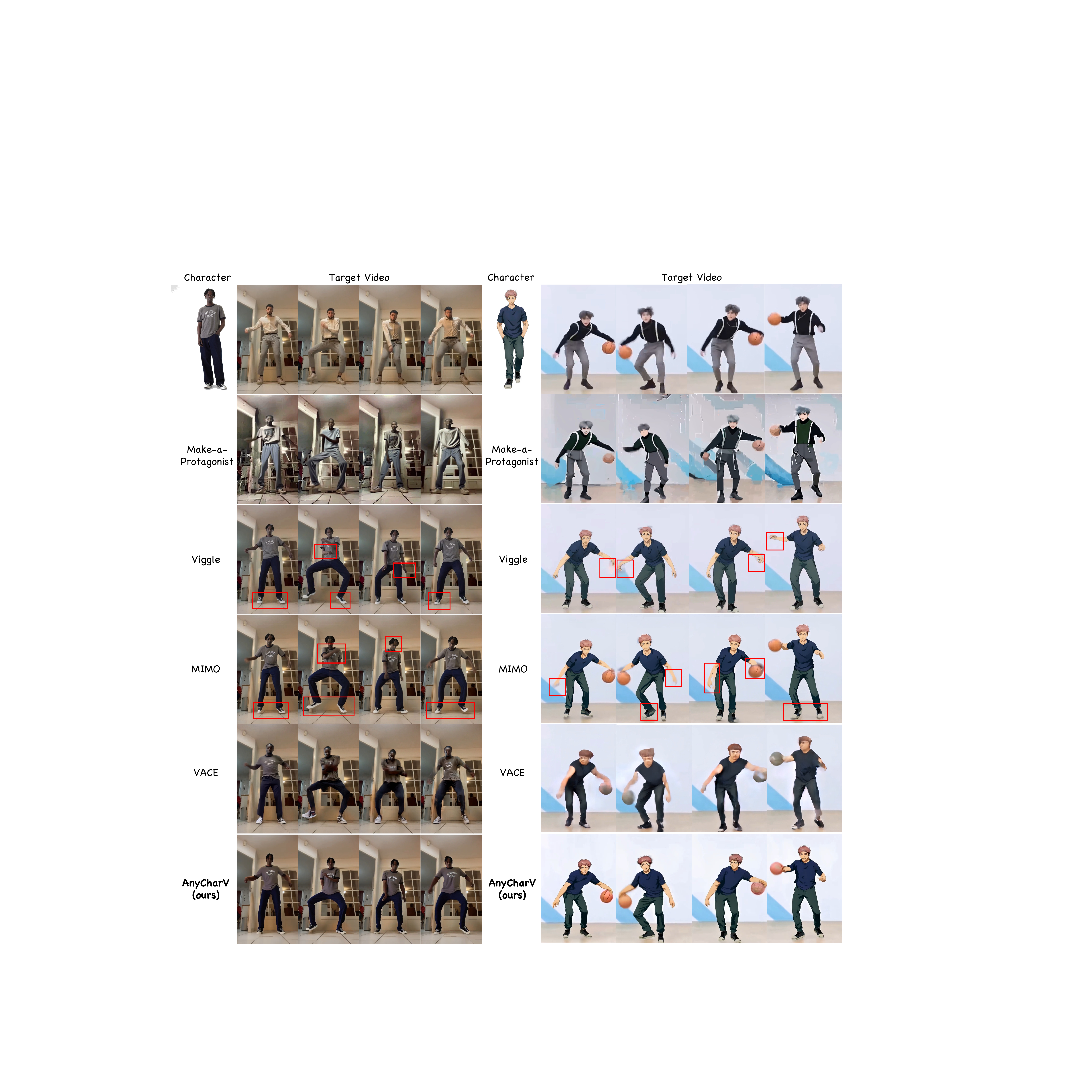}
    \vspace{-7mm}
    \caption{Qualitative results of our method compared with previous SOTA methods. The reference character and target video are shown in the top. Each following line indicates a method.}
    \label{fig:comparison_sota}
\end{figure*}

\subsection{Training and Inference}

\paragraph{Training Strategy.}
The training of our model is carried out in a two-stage manner.
In the first stage, we train the denoising UNet, ReferenceNet, and pose guider and fix VAE \cite{kingma2013auto} and CLIP image encoder \cite{radford2021learning}.
The training objective is as following:
\begin{equation}
\mathcal{L}_{\text{1}}=\mathbb{E}_{\epsilon, \boldsymbol{z}, \boldsymbol{x}_r, \boldsymbol{s}_d, \boldsymbol{m}_d, \boldsymbol{p}_d, t}\left[\left\|\epsilon\!-\!\epsilon_{\Theta}\left(\boldsymbol{z}_t, \boldsymbol{x}_r, \boldsymbol{s}_d, \boldsymbol{m}_d, \boldsymbol{p}_d, t\right)\right\|_2^2\right].
\label{equ:first_stage}
\end{equation}
In the second stage, we load the trained weights of the denoising UNet, ReferenceNet, and pose guider in the first stage and only finetune the denoising UNet to boost the capability of our generation model. 
The training loss during this stage is
\begin{equation}
\mathcal{L}_{\text{2}}=\mathbb{E}_{\epsilon, \boldsymbol{z}', \boldsymbol{x}_r^d, \boldsymbol{x}_d^r, \boldsymbol{m}_d^{c\prime}, \boldsymbol{p}_d', t}\left[\left\|\epsilon\!-\!\epsilon_{\Theta}\left(\boldsymbol{z}_{t}', \boldsymbol{x}_r^d, \boldsymbol{x}_d^r, \boldsymbol{m}_d^{c\prime}, \boldsymbol{p}_d', t\right)\right\|_2^2\right],
\label{equ:second_stage}
\end{equation}
where $\epsilon$ is the randomly sampled noise, $\epsilon_\Theta$ is the noise prediction from $\Theta$, $\boldsymbol{z}'$ is the latent of the target driving video $\boldsymbol{x}_d'$, $\boldsymbol{p}_d'$ is the 2D pose sequence extracted from reference driving video $\boldsymbol{x}_d^r$, and $t$ is the sampling timestep.

\paragraph{Inference.}
During inference, we only need a reference image and a target driving video to generate the desired output. 
The target character's pose and bounding box mask can be extracted in advance.
With our proposed two-stage training mechanism, our model can produce high-fidelity videos that accurately preserve the identity of the reference character and the target background scene. 
The generated videos exhibit smooth motions and natural interactions.


\section{Experiment}


\subsection{Experimental Setting}
\label{sec:exp_setting}

We build a character video dataset called CharVG to train our proposed model, which contains 9,055 videos from the Internet with various characters.
These videos range from 7 to 59 seconds in length.
The ReferenceNet and denoising UNet are both initialized from Stable Diffusion 1.5 \cite{rombach2022high}, where the motion module in the denoising UNet is initialized from AnimateDiff \cite{guo2024animatediff}.
The pose guider is initialized from ControlNet \cite{zhang2023adding}.
We conduct both qualitative and quantitative evaluation for our method.
For quantitative evaluation, we collect 20 character images and 10 target driving videos from the internet, then generate videos with every image-video pair, which results in 200 evaluation videos.
We adopt FVD \cite{unterthiner2018towards}, Dover++ \cite{wu2023exploring}, and CLIP Image Score \cite{radford2021learning} to evaluate the generation quality.
Please refer to Section \ref{sec:appendix_exp_setting} for detailed experimental settings.


\begin{table}[t]
\begin{minipage}{0.47\linewidth}
\centering

\makeatletter\def\@captype{table}\makeatother

  \resizebox{\linewidth}{!}{%

\begin{tabular}{lccc}
\toprule
Method & FVD $\downarrow$ & DOVER++$\uparrow$ & CLIP-I$\uparrow$ \\
\midrule
MAP \cite{zhao2023make} & 671.44  & 34.91  & 66.03  \\
Viggle \cite{viggle2024} & 638.92  & 55.47  & 67.97  \\
MIMO \cite{men2024mimo} & 647.12  & 54.89  & 68.01  \\
VACE \cite{jiang2025vace} & 658.56  & 55.26  & 67.93  \\
\midrule
\rowcolor[rgb]{ 1,  .882,  .8} \textbf{AnyCharV (ours)} & \textbf{622.01 } & \textbf{57.07 } & \textbf{69.14 } \\
\bottomrule
\end{tabular}%

}

\caption{Quantitative results of our proposed method AnyCharV compared with previous SOTA methods. MAP denotes Make-A-Protagonist. The best results are in bold.} 
\tabcolsep=1.3mm
\label{tab:comparison_sota}

 \end{minipage}
 \hfill
\begin{minipage}{0.51\linewidth}
\centering

\makeatletter\def\@captype{table}\makeatother

  \resizebox{\linewidth}{!}{%

\begin{tabular}{lccc}
\toprule
Method & Identity $\downarrow$ & Motion $\downarrow$ & Scene $\downarrow$ \\
\midrule
MAP \cite{zhao2023make} & 4.78$\pm$0.05 & 4.91$\pm$0.07 & 4.99$\pm$0.05 \\
Viggle \cite{viggle2024} & 2.04$\pm$0.24 & 2.20$\pm$0.33 & 2.04$\pm$0.31 \\
MIMO \cite{men2024mimo} & 2.39$\pm$0.26 & 2.12$\pm$0.26 & 2.30$\pm$0.27 \\
VACE \cite{jiang2025vace} & 3.90$\pm$0.34 & 3.75$\pm$0.31 & 3.68$\pm$0.28 \\
\midrule
\rowcolor[rgb]{ 1,  .882,  .8} \textbf{AnyCharV (ours)} & \boldmath{}\textbf{1.89$\pm$0.24}\unboldmath{} & \boldmath{}\textbf{2.02$\pm$0.37}\unboldmath{} & \boldmath{}\textbf{1.99$\pm$0.27}\unboldmath{} \\
\bottomrule
\end{tabular}%

}

\caption{User study results of our proposed method AnyCharV compared with previous SOTA methods. MAP denotes Make-A-Protagonist. The best results are in bold. 
  }
  \tabcolsep=1.3mm
   \label{tab:user_study}%
	
 \end{minipage}

\end{table}

\begin{figure}[t]

\centering

\begin{minipage}[c]{0.505\textwidth}
    \centering
    \includegraphics[width=\textwidth]{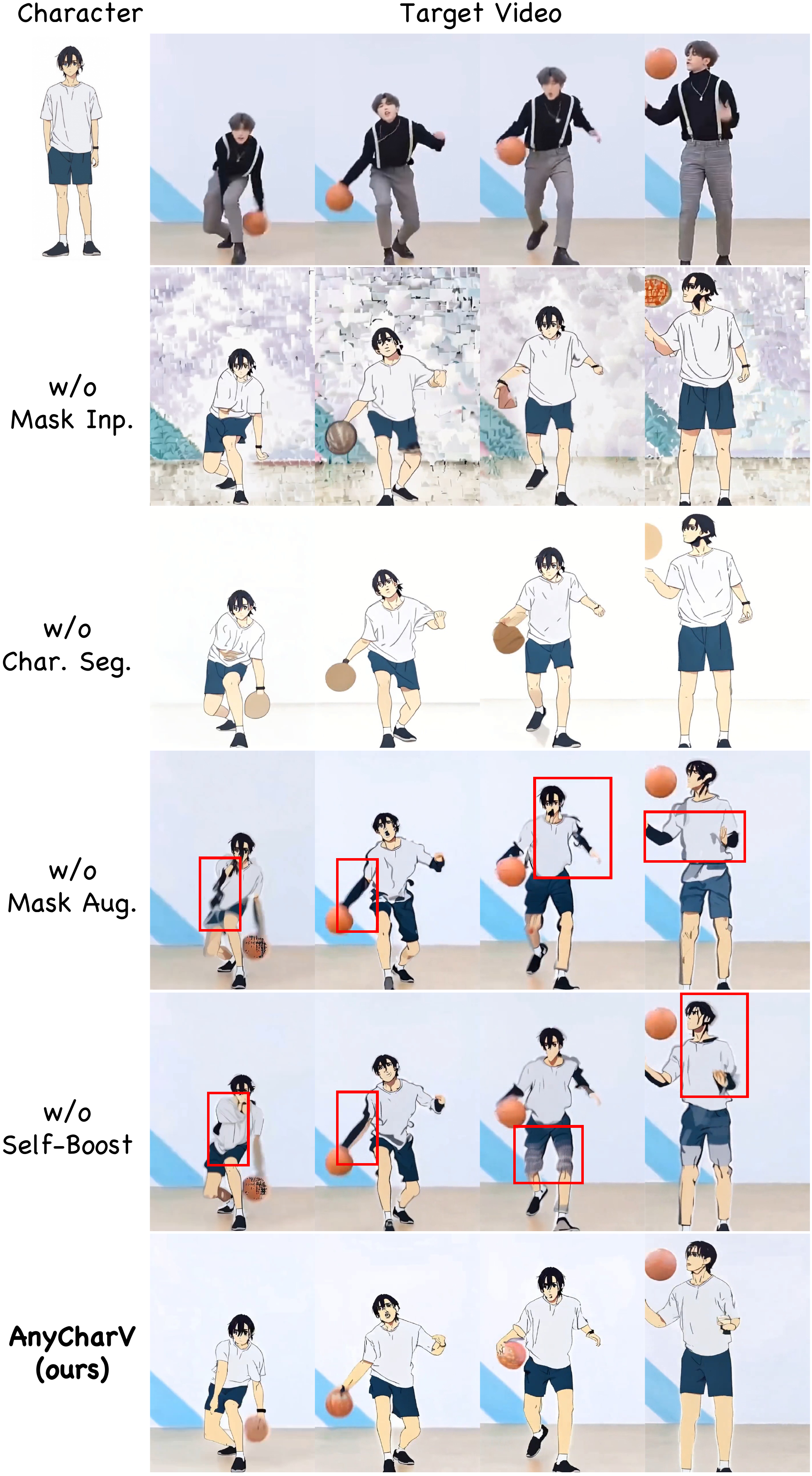}
    \vspace{-7mm}
    \caption{Visualization for the effect of different components. The reference character and target video are shown in the top. Each following line indicates the generated video from a variant.}
    \label{fig:component}
\end{minipage} 
\hfill
\begin{minipage}[c]{0.475\textwidth}
    \centering
    \includegraphics[width=\linewidth]{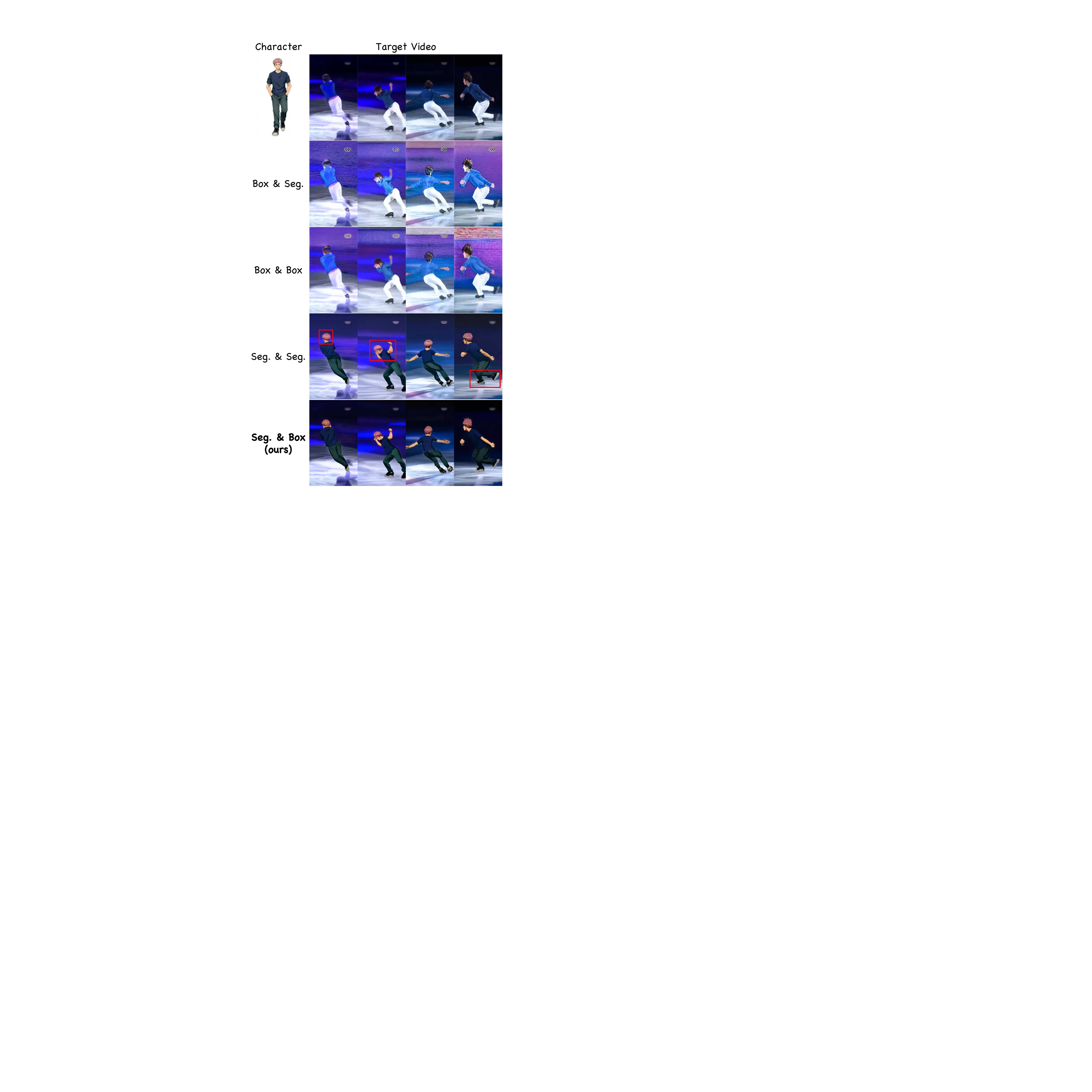}
    \vspace{-7mm}
    \caption{Visualization for the effect of different mask types. The reference character and target video are shown in the top. Each following line indicates the generated video from a variant. `Box \& Seg.' indicates that the  bounding box mask is used in the first stage and the segmentation mask is used in the second stage.}
    \label{fig:mask_type}
\end{minipage}
    
\end{figure}

\begin{table}[t]
\begin{minipage}{0.49\linewidth}
\centering

\makeatletter\def\@captype{table}\makeatother

  \resizebox{\linewidth}{!}{%

\begin{tabular}{lccc}
\toprule
Method & FVD $\downarrow$ & DOVER++ $\uparrow$ & CLIP-I $\uparrow$ \\
\midrule
w/o mask inp. & 680.31  & 54.09  & 66.12  \\
w/o char. seg. & 678.51  & 55.30  & 67.97  \\
w/o mask aug. & 646.03  & 55.58  & 68.72  \\
w/o self-boost & 652.34  & 53.54  & 67.75  \\
\midrule
\rowcolor[rgb]{ 1,  .882,  .8} \textbf{AnyCharV (ours)} & \textbf{622.01} & \textbf{57.07} & \textbf{69.14} \\
\bottomrule
\end{tabular}%

}

\caption{The effect of different components. We can learn that every design contributes to the superior performance.} 
\label{tab:component}
\tabcolsep=1.3mm
	
 \end{minipage}
 \hfill
\begin{minipage}{0.49\linewidth}
\centering

\makeatletter\def\@captype{table}\makeatother

  \resizebox{\linewidth}{!}{%

\begin{tabular}{lccc}
\toprule
Mask Type & FVD $\downarrow$ & DOVER++$\uparrow$ & CLIP-I$\uparrow$ \\
\midrule
Box \& Seg. & 702.16 & 54.64 & 66.47 \\
Box \& Box & 670.86 & 54.23 & 67.46 \\
Seg. \& Seg. & 648.61 & 55.84 & 68.07 \\
\rowcolor[rgb]{ 1,  .882,  .8} \textbf{Seg. \& Box (ours)} & \textbf{622.01} & \textbf{57.07} & \textbf{69.14} \\
\bottomrule
\end{tabular}%

}

\caption{The effect of different mask types. `Box \& Seg.' indicates that the  bounding box mask is used in the first stage and the segmentation mask is used in the second stage. 
  }
  \tabcolsep=1.3mm
   \label{tab:mask_type}%
	
 \end{minipage}

\end{table}



\begin{figure}[t]
    \centering
    \includegraphics[width=\linewidth]{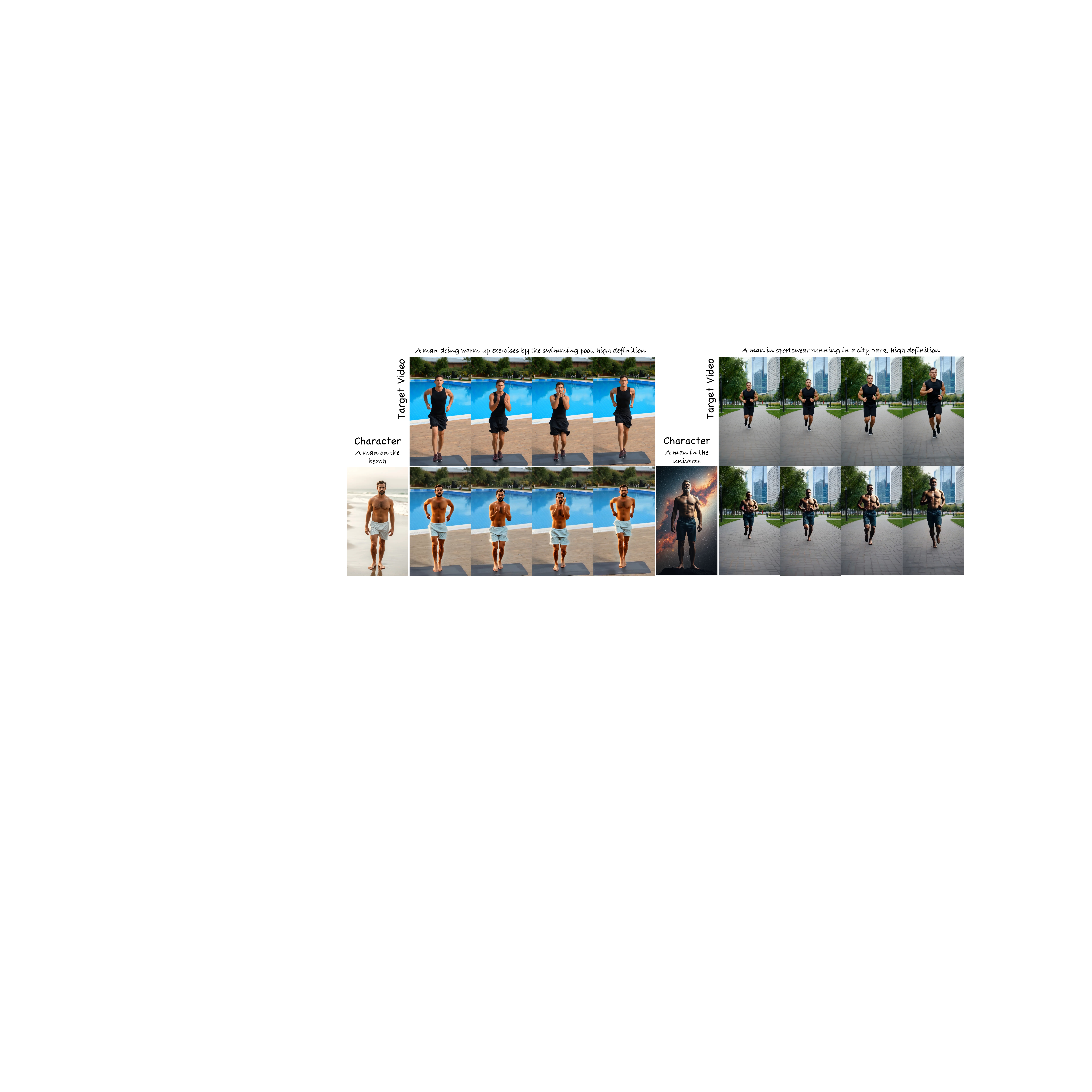}
    \vspace{-7mm}
    \caption{Qualitative results of combining AnyCharV with FLUX \cite{flux2023} and HunyuanVideo \cite{kong2024hunyuanvideo}. The text prompts used for generating reference image and target video are given above them, respectively.}
    \label{fig:application}
\end{figure}

\subsection{Comparison}

We compare our approach with four recent state-of-the-art methods, including Make-A-Protagonist \cite{zhao2023make}, Viggle \cite{viggle2024}, MIMO \cite{men2024mimo}, and VACE \cite{jiang2025vace}.


\paragraph{Qualitative Results.}

As shown in Figure \ref{fig:comparison_sota}, Make-a-Protagonist performs worst as for lacking effective spatial and temporal modeling.
Notably, our AnyCharV can preserve more detailed appearance and avoid lots of artifacts than Viggle and MIMO, especially looking at the generated arms and hands in Figure \ref{fig:comparison_sota}.
VACE can not preserve the identity of the reference character very well.
Moreover, our approach can handle the complex human-object interactions very well, {\it e.g.}, playing basketball, which can not be done with Make-a-Protagonist, Viggle, and VACE.
These results strongly affirm the effectiveness and robustness of our proposed AnyCharV.


\paragraph{Quantitative Results.}
We show the results in Table \ref{tab:comparison_sota}. 
Our AnyCharV surpasses the best open-sourced model VACE by a significant margin, with 5.55\% FVD, 3.17\% DOVER++, and 1.78\% CLIP Image Score improvements.
Moreover, AnyCharV performs much better than Viggle and MIMO, which are both closed-source industrial products.
For example, AnyCharV outperforms MIMO by 3.88\%, 3.97\%, 1.66\% in terms of FVD, DOVER++, and CLIP Image Score.
These results clearly demonstrate the great effectiveness and robustness comparing with such closed-source industrial product.
For inference cost, Make-A-Protagonist, Viggle, MIMO, VACE, and AnyCharV take 140, 2, 8, 10, and 5 minutes to generate a 5 seconds 24 FPS video with resolution $576 \times 1024$, respectively, further denoting the high efficiency of our approach.

\paragraph{User Study.}
We conduct human evaluations to further evaluate our approach, comparing with four SOTA methods.
We gather 1,200 answers from 30 independent human raters, evaluating the identity preservation of the reference character, the motion consistency between the generated video and the target driving video, and the scene similarity and interaction.
The average ranking is computed and shown in Table \ref{tab:user_study}.
This shows that our AnyCharV significantly outperforms the open-source Make-A-Protagonist and MACE as well as the closed-source models Viggle and MIMO, demonstrating the effectiveness of our model.
More details of the metrics can be found in Section \ref{sec:app_eval}.

\subsection{Ablation Studies}


\paragraph{Mask Input.}

For both two stages, we provide the mask information to help the model locate the area of the character and further composition. From the results in from Table \ref{tab:component} and Figure \ref{fig:component}, we find that without the mask input, the model fails to compose the source character and target scene. 
These results also demonstrate that masking the character in the target video can not incorporate fine segmentation information in the first stage to guide the composition.

\paragraph{Character Segmentation.}
During training, we segment the character out for all reference images to avoid introducing redundant context information which can degrade the quality of the generated videos. From Table \ref{tab:component} and Figure \ref{fig:component}, we can observe that the unnecessary background in the reference character will dramatically degrade the quality of the generated character. These results validate the necessity for the segmentation of characters.

\paragraph{Mask Augmentation.}
AnyCharV augments the segmentation mask during the first stage to reduce the negative effect caused by the fine mask shape.
The qualitative and quantitative results shown in Table \ref{tab:component} and Figure \ref{fig:component} demonstrate the effectiveness of our design.
Without mask augmentation, the appearance of the reference character is disturbed greatly.

\paragraph{Self-Boosting Strategy.}
We introduce a self-boosting training mechanism to enhance the identity preservation.
From Figure \ref{fig:component}, we can learn that without self-boosting training, the identity and appearance of the reference character can not be well preserved, especially for the dressing cloth and face. 
Such performance drop can also be observed in Table \ref{tab:component}, indicating the effectiveness of our self-boosting training.

\paragraph{Mask Type.}
We conduct ablation on different mask types during two stages.
In the first training stage, the model composes the reference character and the target video scene within the accurate region, where the fine segmentation mask is preferred, indicated by the great spatial information loss for both character and background scene, produced by lines `Box \& Seg.' and `Box \& Box' in Figure \ref{fig:mask_type} and Table \ref{tab:mask_type}.
In the second training stage, the model is expected to better preserve the details of the reference character.
In this case, we use more loose mask constrain, {\it i.e.}, coarse bounding box mask, to guide the character generation, eliminating the adverse effect caused by the fine segmentation mask shape.
Such performance improvement can be validated in Table \ref{tab:mask_type} and Figure \ref{fig:mask_type}.

\subsection{Application}

Our AnyCharV can also be used to generate a video with a reference image generated by text-to-image (T2I) models, {\it e.g.}, FLUX \cite{flux2023}, driving by a target video generated by text-to-video (T2V) models, {\it e.g.}, HunyuanVideo \cite{kong2024hunyuanvideo}.
As shown in Figure \ref{fig:application}, we first utilize FLUX to generate a reference image and HunyuanVideo to synthesize a target video, then the generated target video is used to drive the synthesized reference character.
These results clearly indicate the strength and flexibility of our AnyCharV, proving its versatility.

\section{Conclusion}

In this work, we introduce a novel framework \textbf{\textit{AnyCharV}} with fine-to-coarse guidance for controllable character video generation under a two-stage training strategy.
The self-supervised composition strategy with fine mask guidance in the first stage basically learn to drive a reference image with a target driving video to guarantee the motion correctness and target scene maintenance.
Self-boosting training is then carried out via building interactions between the reference and target characters with coarse mask guidance, where the detailed identity of the reference character can be better preserved.
AnyCharV clearly beats both SOTA open-source models and leading closed-source industrial products.
Most importantly, AnyCharV can be used for images and videos created by T2I and T2V models, showing its strong ability to generalize.

{
\small
\bibliographystyle{unsrt}
\bibliography{neurips_2025}
}

\clearpage
\appendix

{\bf Roadmap.}
The appendix is organized as the following: 
Section \ref{sec:appendix_exp_setting} describes the detailed experimental settings;
Section \ref{sec:appendix_exp_results} shows additional experimental results;
Section \ref{sec:appendix_limitations} discusses some possible limitations and our future works;
Section \ref{sec:broader_impact} demonstrates the potential broader impacts of our work.

\section{Detailed Experimental Setting}
\label{sec:appendix_exp_setting}

\subsection{Dataset}
We build a character video dataset called CharVG to train our proposed model, which contains 9,055 videos from the internet with various characters.
These videos range from 7 to 59 seconds in length.
We utilize DWPose \cite{yang2023effective} to extract the whole-body poses from these videos, and the estimated poses are rendered as pose skeleton image sequences.
Meanwhile, we use YOLOv8 \cite{reis2023real} and SAM2 \cite{ravi2024sam} to obtain the segmentation and bounding box masks for the characters presented in these videos.

\subsection{Implementation Details}
The ReferenceNet and denoising UNet are both initialized from Stable Diffusion 1.5 \cite{rombach2022high}, where the motion module in the denoising UNet is initialized from AnimateDiff \cite{guo2024animatediff}.
The pose guider is initialized from ControlNet \cite{zhang2023adding}.
In our training, firstly, we train the ReferenceNet, denoising UNet without motion module, and pose guider with individual frames from videos for 50,000 steps to learn the spatial information.
Secondly, we perform self-boosted training on only denoising UNet without motion module with individual video frames for 3,000 steps to help the model eliminate the bad influence of the mask shape.
Above two training processes are both conducted with resolution $768 \times 768$ and batch size 64.
Thirdly, we only train the motion module in the denoising UNet with 24-frame video clips with resolution $768 \times 768$ for 10,000 steps to improve the temporal consistency.
Last, the self-boosted training strategy is performed only on the denoising UNet with resolution $704 \times 704$ for 10,000 steps to better preserve the identity of the reference character.
Above two training processes are both conducted with batch size 8.
All above are trained using learning rate 1e-5.
During inference, the long video is generated sequentially with several short clips and temporally aggregated following \cite{tseng2023edge}.
We use a DDIM \cite{song2020denoising} scheduler for 30 denoising steps with classifier-free guidance \cite{ho2021classifierfree} as 3.0.
All of our experiments are finished on 8 NVIDIA H800 GPUs using PyTorch \cite{paszke2019pytorch}.
The overall training process takes about 3 days.
Our model takes 5 minutes to generate a 5 seconds 24FPS video with resolution $576 \times 1024$ on a single NVIDIA H800 GPU.

\subsection{Evaluation Metrics}
\label{sec:app_eval}

\paragraph{Quantitative Evaluation.}
We adopt three metrics to evaluate the generation quality:
\begin{itemize}
    \item FVD \cite{unterthiner2018towards} score is a widely used metric for assessing the generated videos. We compute FVD score between our generated videos and 1,000 real character videos from our dataset.
    \item Dover++ \cite{wu2023exploring} score is a video quality assessment metric from both aesthetic and technical perspective, demonstrating the overall quality of the generated videos.
    \item CLIP Image Score \cite{radford2021learning} is used to evaluate the similarity between the generated video and the reference character, validating the model capability for preserving the identity.
\end{itemize}

\paragraph{User Study.}
In user study, the quality of the generated video are assessed from three aspects:
\begin{itemize}
    \item Character identity preservation (Identity in Table \ref{tab:comparison_sota}): this metric is used to evaluate whether the reference character remains unchanged in the generated videos.
    \item Motion consistency (Motion in Table \ref{tab:comparison_sota}): this metric measures the fluidity of the character’s motion and whether the motions match between the target video and the generated video.
    \item Scene similarity/interaction (Scene in Table \ref{tab:comparison_sota}): this metric assesses the preservation of the background scene shown in the target video. Also paying attention to the interaction, as the object is also a part of the background scene.
\end{itemize}

We have tried our best to carry out the user study under a diverse scenario from two aspects:
\begin{itemize}
    \item Various samples: we generate various videos with different types of target scenes (sports, dancing, film, …) and different reference characters (real human, anime character, AI generated character, …). 
    \item Diverse human raters: the generated videos are evaluated by human raters with diverse backgrounds, including relevant researchers, financial practitioners, teachers, and so on.
\end{itemize}

\subsection{Comparison Methods}

We compare our proposed AnyCharV with three state-of-the-art methods. We describe these methods as follows:
\begin{itemize}
    \item Make-A-protagonist \cite{zhao2023make} memorizes the visual and motion information of the target video by fine-tuning the video generation model and guiding the generation of the desired output through masks and reference images during the inference stage.
    \item Viggle \cite{viggle2024} performs rapid 3D modeling of the reference image and manipulates it according to the pose sequence of the input video, thereby accomplishing tasks such as character generation.
    \item MIMO \cite{men2024mimo} decomposes the character's identity, pose, and scene to facilitate video synthesis. And it can be used for composing the character and scene.
    \item VACE \cite{jiang2025vace} learns various video generation and editing tasks in a unified manner, which can be applied for character replacement.
\end{itemize}

\section{Additional Experimental Results}
\label{sec:appendix_exp_results}

\begin{figure*}[!htb]
    \centering
    \includegraphics[width=\linewidth]{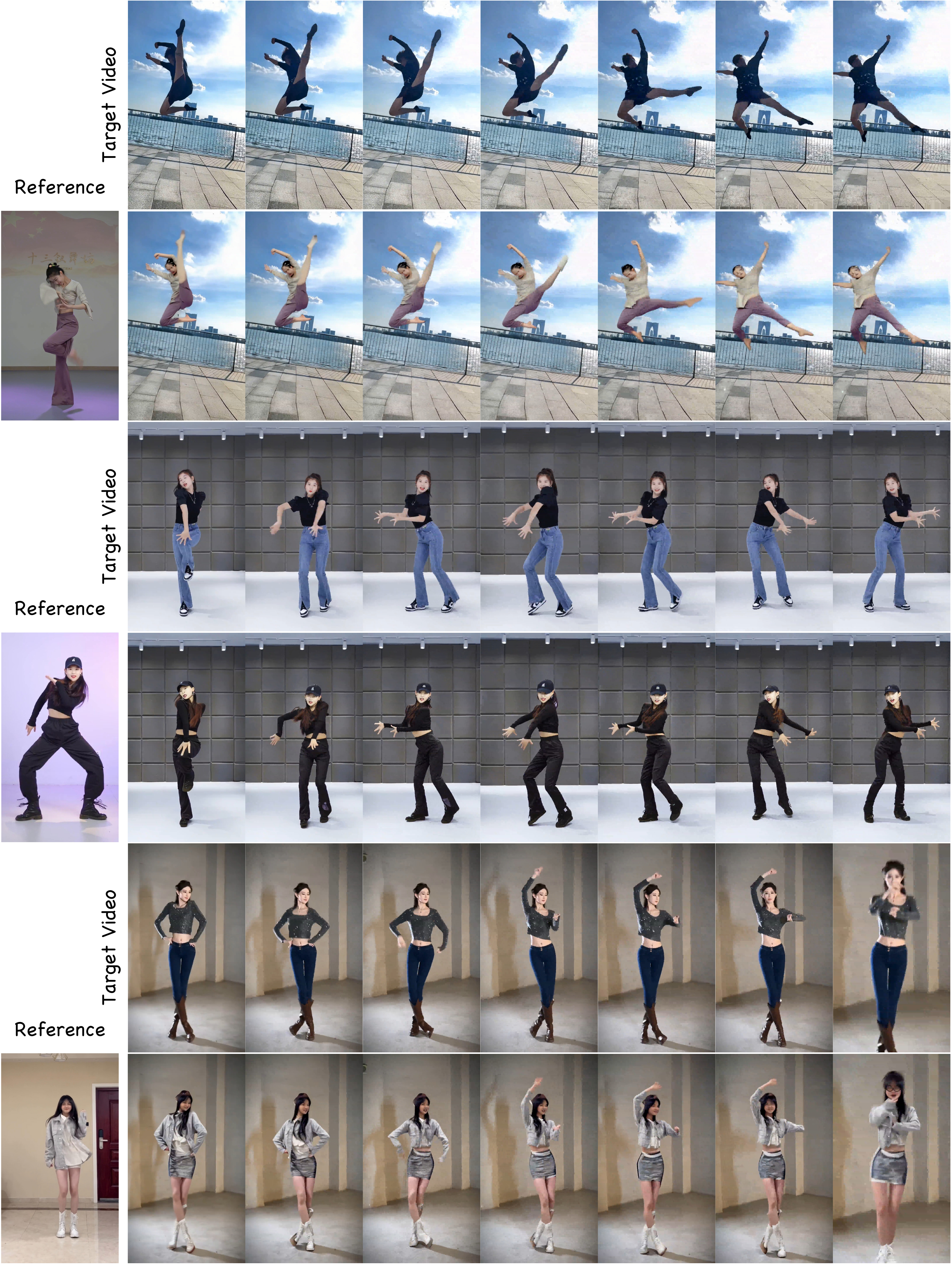}
    \caption{Qualitative visualization results of our generated data pairs with a reference image (left) and a target video (top).}
    \label{fig:generated_pair}
\end{figure*}

\paragraph{Synthesized Video Pairs.}
We build 64,000 video pairs for training in the second stage. Here, we illustrate some synthesized video pairs in Figure \ref{fig:generated_pair}.
We could learn that the high-quality synthesized video pairs strongly support the effective learning in the second stage.

\begin{figure*}[!htb]
    \centering
    \includegraphics[width=\linewidth]{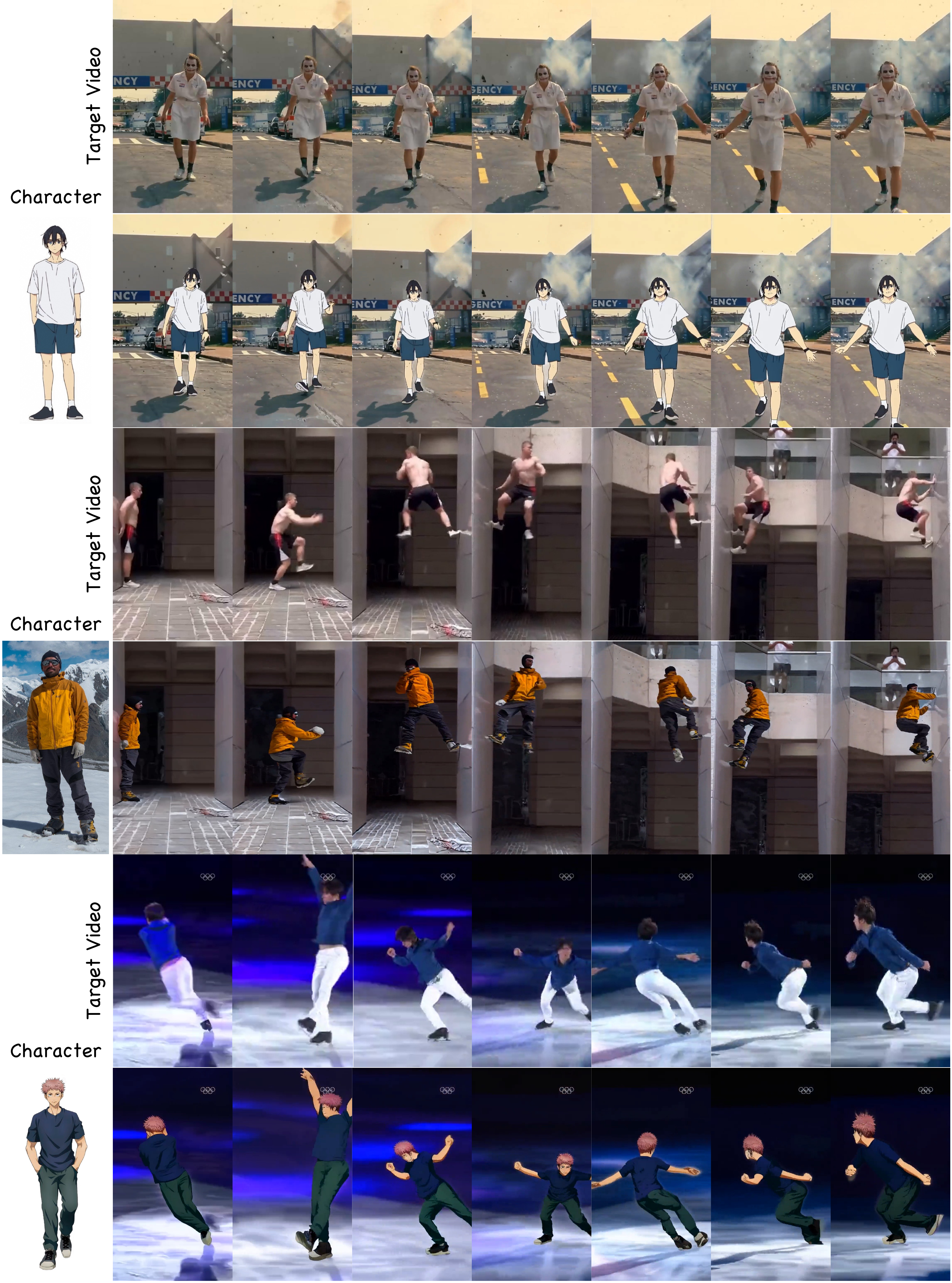}
    \caption{Qualitative visualization results of our method given a reference image (left) and a target video (top).}
    \label{fig:visualization1}
\end{figure*}

\begin{figure*}[!htb]
    \centering
    \includegraphics[width=\linewidth]{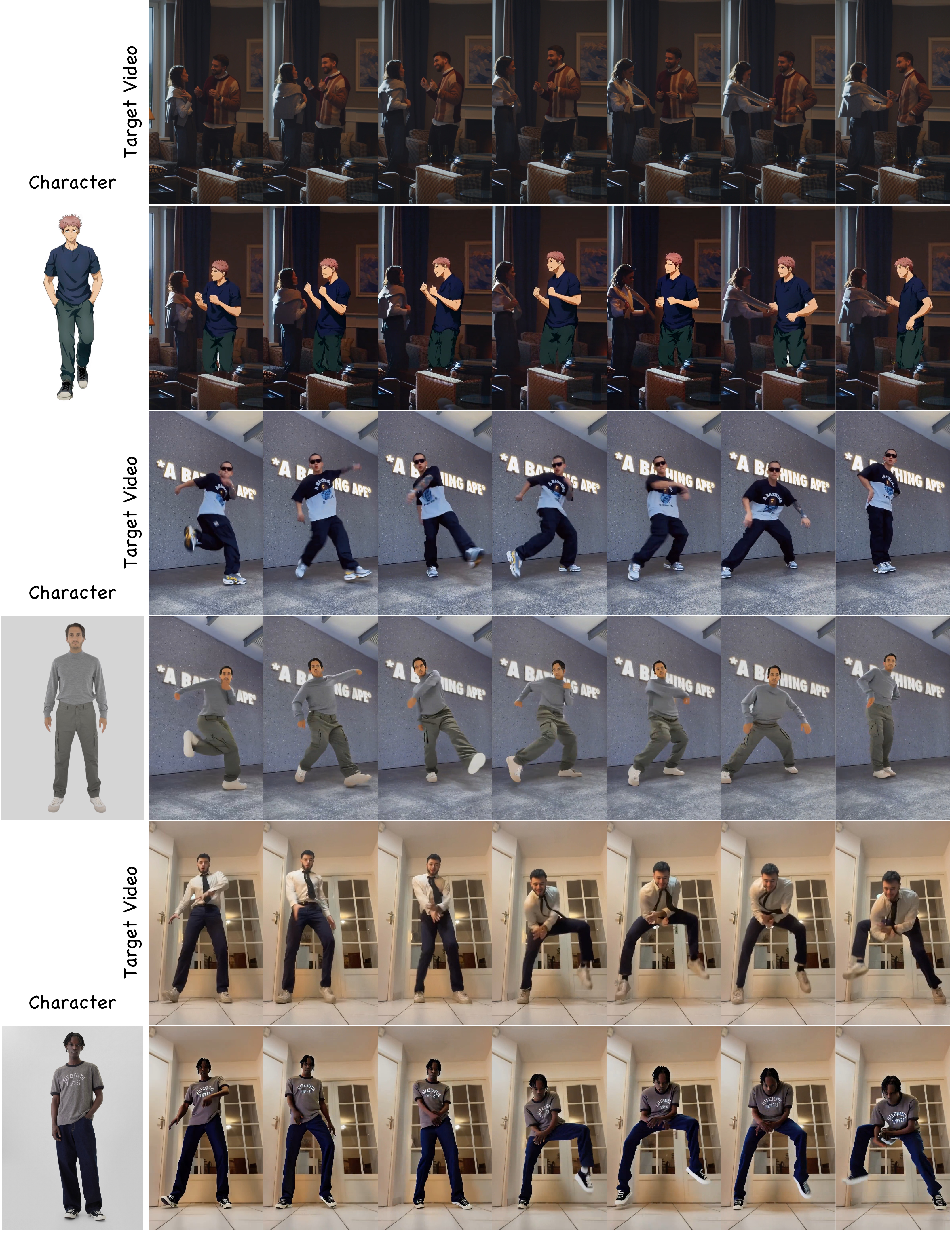}
    \caption{Qualitative visualization results of our method given a reference image (left) and a target video (top).}
    \label{fig:visualization2}
\end{figure*}

\paragraph{Qualitative Results.}
We show more visualization results with diverse characters, scenes, motions, and human-object interactions in Figure \ref{fig:visualization1} and Figure \ref{fig:visualization2}, further demonstrating the robustness and effectiveness of our proposed method.

\paragraph{User Interface.}

We design a user interface to help human raters easily evaluate the generated videos. 
Our designed user interface is illustrated in Figure \ref{fig:user_interface}.

\paragraph{Demo Video.}

We attach a demo video in the supplementary material, including three parts: 1) comparison with SOTA methods; 2) generated videos with our AnyCharV; 3) application: combining with 
Text-to-Image and Text-to-Video models.
To be note that MIMO automatically intercepts the first half in the first comparison video and VACE automatically sample a subset of frames for generation.

\section{Limitations and Future Works}

\label{sec:appendix_limitations}

While our AnyCharV model demonstrates impressive results in controllable character video generation, it may struggle when tasked with inferring the back view of a character from a front-facing reference image. 
Looking ahead, we plan to enhance the generalization capabilities of AnyCharV by integrating a more robust video generator and introducing more specialized controls for specific scenarios.

\section{Broader Impact}
\label{sec:broader_impact}
Flexibly controlling the generation of different characters is crucial in our practical life.
We achieve composing an arbitrary character and an arbitrary target scene with our proposed two-stage framework AnyCharV.
We strongly believe that our work will be beneficial for a number of AIGC applications, such as creative AI and film production.
However, our model still suffers from generating the back view of the provided character.
Please be careful to use our model when meeting such scenario.

\begin{figure*}[!htb]
    \centering
    \includegraphics[width=0.85\linewidth]{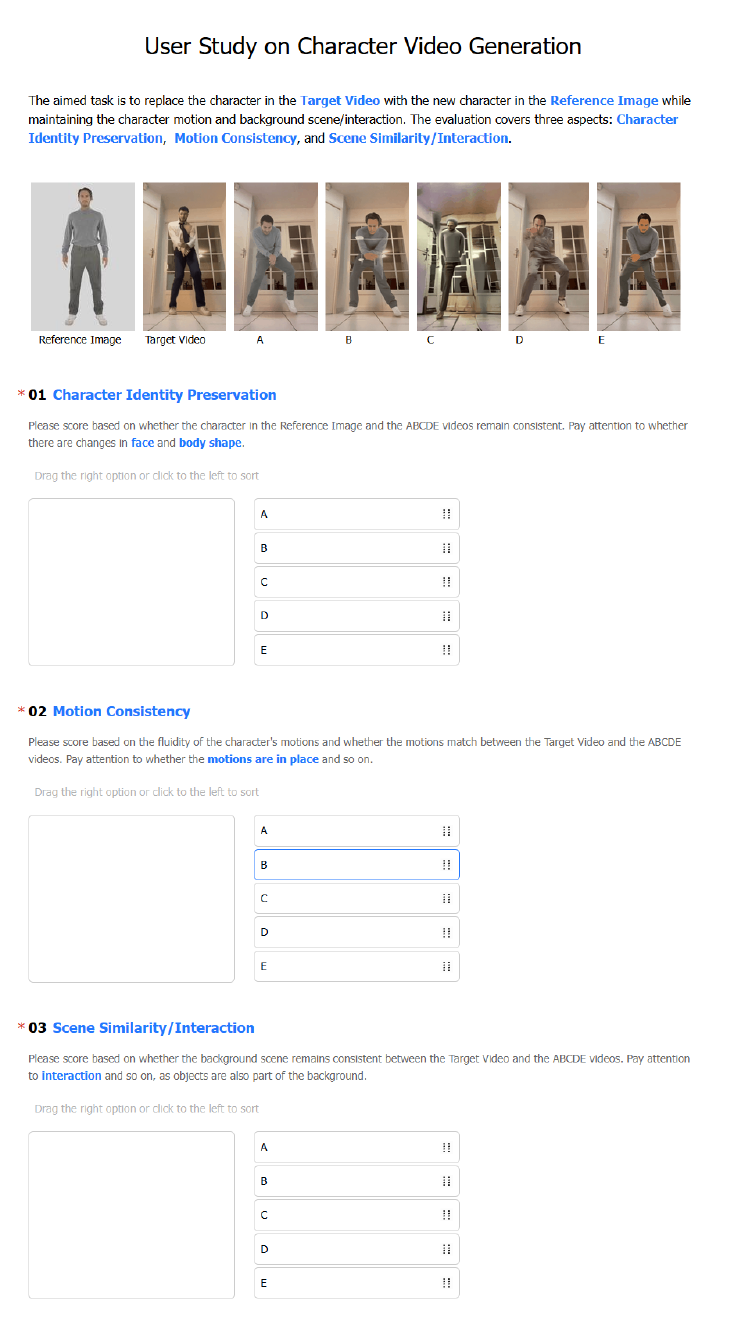}
    \caption{Our user interface for user study.}
    \label{fig:user_interface}
\end{figure*}

\end{document}